\documentclass[10pt,twocolumn,letterpaper]{article}

\usepackage[pagenumbers]{iccv} 

\usepackage[pagebackref,breaklinks,colorlinks,allcolors=iccvblue]{hyperref}
\usepackage{afterpage}
\usepackage{stfloats}

\usepackage{multirow}
\usepackage{multicol}
\usepackage{pifont}
\usepackage{booktabs}
\usepackage{amssymb}

\usepackage{colortbl}  
\newcommand{\cmark}{\ding{51}}%
\newcommand{\xmark}{\ding{55}}%
\usepackage{threeparttable}
\usepackage{subcaption} 
\usepackage{color}
\definecolor{iccvblue}{rgb}{0.21,0.49,0.74}
\definecolor{mygreen}{RGB}{93,174,86}
\definecolor{mygray}{gray}{0.9}

\def\confName{ICCV}
\def\confYear{2025}
\usepackage{marvosym} 
\title{MultiCo3D: Multi-Label Voxel Contrast for One-Shot Incremental Segmentation of 3D Neuroimages
}

\author{
\vspace{-10mm}\\
Hao Xu$^{1}$, Tengfei Xue$^{1}$, Dongnan Liu$^{1}$, Yuqian Chen$^{2,4}$, Fan Zhang$^{3}$, \\
Carl-Fredrik Westin$^{2,4}$, Ron Kikinis$^{2,4}$, Lauren J. O'Donnell$^{2,4}$, Weidong Cai$^{1}$$^($\textsuperscript{\Letter}$^)$\\
$^{1}$The University of Sydney, Sydney, Australia\\
$^{2}$Harvard Medical School, Boston, USA\\
$^{3}$University of Electronic Science and Technology of China, Chengdu, China\\
$^{4}$Brigham and Women's Hospital, Boston, USA\\
tom.cai@sydney.edu.au
\vspace{-5mm}
}

\begin{document}
\maketitle
\begin{abstract}
3D neuroimages provide a comprehensive view of brain structure and function, aiding in precise localization and functional connectivity analysis. Segmentation of white matter (WM) tracts using 3D neuroimages is vital for understanding the brain's structural connectivity in both healthy and diseased states. 
One-shot Class Incremental Semantic Segmentation (OCIS) refers to effectively segmenting new (novel) classes using only a single sample while retaining knowledge of old (base) classes without forgetting. Voxel-contrastive OCIS methods adjust the feature space to alleviate the feature overlap problem between the base and novel classes. However, since WM tract segmentation is a multi-label segmentation task, existing single-label voxel contrastive-based methods may cause inherent contradictions. To address this, we propose a new multi-label voxel contrast framework called MultiCo3D for one-shot class incremental tract segmentation. Our method utilizes uncertainty distillation to preserve base tract segmentation knowledge while adjusting the feature space with multi-label voxel contrast to alleviate feature overlap when learning novel tracts and dynamically weighting multi losses to balance overall loss. We compare our method against several state-of-the-art (SOTA) approaches. The experimental results show that our method significantly enhances one-shot class incremental tract segmentation accuracy across five different experimental setups on HCP and Preto datasets.
\vspace{-7mm}
\end{abstract}    
\section{Introduction}
\label{sec:intro}

\begin{figure}[t]
  \centering
   \includegraphics[width=0.9\linewidth]{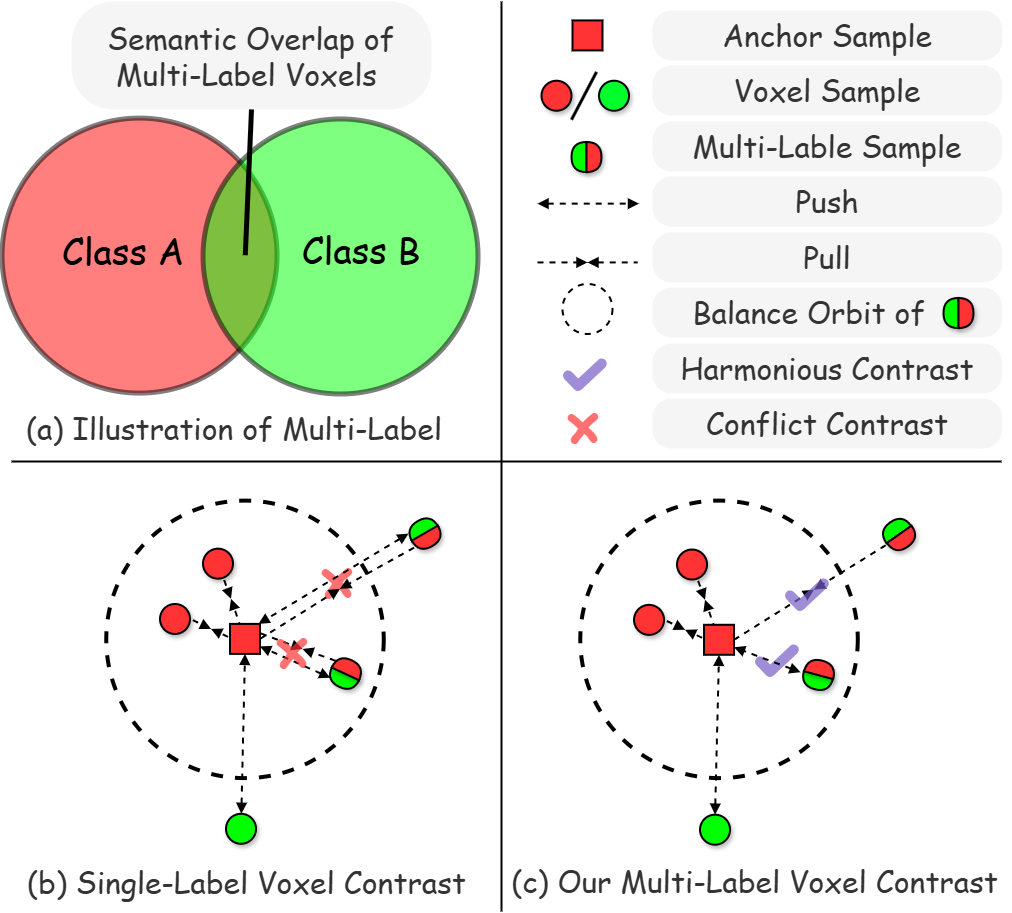}
   \vspace{-3mm}
   \caption{The Conflict of Single-Label Voxel Contrast and Our Proposed Multi-Label Voxel Contrast method. (a) Multi-Label samples have semantic overlap. (b) The conflict of the current single-label voxel contrast method, e.g., \cite{CoinSeg}, may mistakenly pull and push the same sample simultaneously due to the binary nature of positive and negative sample distinctions. (c) Rather than relying on binary positive-negative sample distinctions, our multi-label voxel contrast method adjusts feature space based on their label similarity. This allows us to pull semantically overlapping samples closer and avoid overly pushing or pulling them apart. The balance orbit represents the stable region of semantically overlapping voxels, calculated based on the label similarity between the anchor and the voxel itself. Multi-Label voxels on the orbit remain stable (neither pushed away nor pulled closer).}
   \label{fig:1}
   \vspace{-4mm}
\end{figure}

\begin{figure*}[t]
  \centering
   \includegraphics[width=0.98\linewidth]{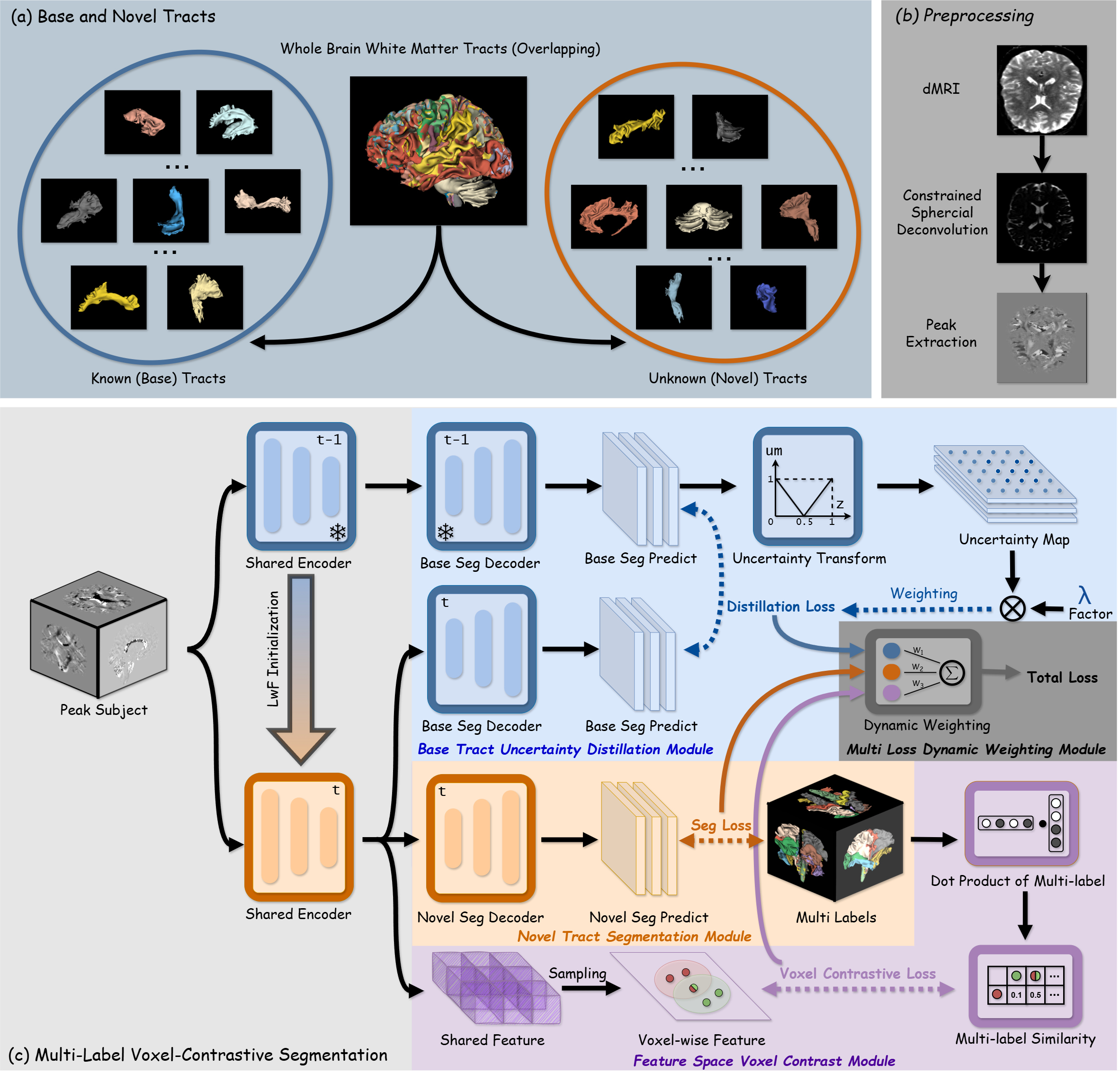}
   \vspace{-0.2cm}
   \caption{The overview of our proposed method. (a) The known (base) and unknown (novel) tracts. (b) The 3D peaks are extracted from dMRI data after preprocessing. (c) Our proposed multi-label voxel-contrast framework.}
   \label{fig:2}
   \vspace{-0.5cm}
\end{figure*}

The human brain can be divided into many functional regions \cite{li2021braingnn,kraaijenvanger2020impact,shallice1994brain}, responsible for movement, vision, hearing, language, memory, etc. These brain regions are connected in white matter (WM) tracts, which form corticocortical or cortico-subcortical connections \cite{Zhang2022-qm}. WM tract segmentation is the process of segmenting specific tracts in the brain, which can help humans understand the higher neurological functions of the brain and the causes of many diseases, improve the surgical effect, and reduce the probability of brain injury and complications. Diffusion Magnetic Resonance Imaging (dMRI) tractography \cite{Zhang2020-mg} can represent human brain  WM connections. The dMRI-based WM tract segmentation is important in analyzing WM characteristics in healthy and diseased brains \cite{Zhang2022-qm, Zhang2020-mg,Li2020-nt,TractCloud}.

Brain tract segmentation is a multi-label voxel segmentation task because different tracts may naturally cross or overlap. Streamline-based methods \cite{Astolfi2020-ke,Zhang2022-qm,Zhang2022-tk,Zhang2022-xr,Xue2022-mz,Xue2023-bm,Chen2021-wa} and voxel-based methods \cite{Wasserthal2018-jj, Lu2021-fp,Liu2022-oa,Lu2022-iv, Liu2022-se} have achieved promising segmentation results. However, the acquisition of annotated data is expensive, especially for rare fiber tracts in clinical practice (such as superficial fiber tracts), which usually require a lot of time and expertise to annotate. Therefore, to reduce dependence on large amounts of annotated data, One-shot Class Incremental Segmentation (OCIS) aims to achieve one-shot segmentation of new (novel) fiber tracts using a single sample while retaining the knowledge of the learned old (base) classes. The main challenge of this task is to avoid overlapping features of base and novel categories, especially in the fiber tract multi-label segmentation. The current single-label voxel contrast method (i.e., CoinSeg \cite{CoinSeg}) aligns features by distinguishing between positive and negative samples, usually pulling positive samples closer and pushing negative samples farther away based on whether the samples belong to the same class. However, this method may cause conflicts when the samples have semantic overlap (as shown in Figure \ref{fig:1}(a)). For example, a sample may have multiple labels. In this case, the model may pull the same sample closer and farther away at the same time, resulting in optimization confusion (as shown in Figure \ref{fig:1}(b)).

To address this, we specifically design a multi-label voxel contrast method through inter-class similarity metrics rather than relying solely on binary positive and negative sample division. 
Through the similarity of sample labels, samples with partial semantic overlap are moderately adjusted: when the distance between the sample and the anchor is greater than the relative distance (outside of its balance orbit), it is pulled closer, otherwise it is pushed away (as shown in Figure \ref{fig:1}(c)). This method can model inter-class relationships more finely and avoid extreme feature pushing or pulling in multi-label tasks. Specifically, we propose a novel multi-label voxel contrast framework, MultiCo3D, for one-shot class incremental tract segmentation.
MultiCo3D adopts a classic multi-task architecture with a shared encoder and task-specific decoders which introduces multi-label voxel contrastive learning to adjust the embedding space of shared features to address the feature overlap problem. MultiCo3D includes four key components: a base tract uncertainty distillation module to preserve base tract segmentation knowledge, a novel tract segmentation module to learn new tract information, a feature space multi-label voxel contrast module to refine feature alignment, and a multi-loss dynamic weighting module to balance the overall loss. Additionally, we perform the LwF initialization \cite{Lwf} method to transfer model weights from time step t-1 to time step t.

Our contributions can be summarized as follows:
1) We propose a novel multi-label voxel contrast framework, MultiCo3D, specifically designed for one-shot class incremental tract segmentation;
2) We propose a new multi-label voxel contrast module by introducing the inter-class similarity measure to better adjust the shared feature space. This approach goes beyond the traditional binary positive-negative sample distinction, addressing feature overlap in multi-label novel tract segmentation tasks;
3) We introduce a base tract uncertainty distillation module that better preserves base tract segmentation knowledge;
4) A multi-loss dynamic weighting module is introduced to adaptively balance the total loss, improving segmentation performance.
\section{Related Work}
\subsection{White Matter Tract Segmentation}
Currently, deep-learning-based methods \cite{croitoru2023diffusion,liu2022video,vaswani2017attention,Chen2021-wa,Xue2023-bm,wang2021exploring,zhou2022rethinking} have been widely used for white matter tract segmentation. They can be divided into fiber-based tract segmentation and voxel-based tract segmentation methods. Fiber-based tract segmentation \cite{Xue2022-mz, Zhang2020-mg,Zhang2022-qm} refers to the classification of fiber or streamlines in tractography to form a set of white matter fibers or a tract. These fibers have biological meanings that form a corticocortical or corticosubcortical connection in the brain. Tractography is a computational process that estimates the anatomical trajectories of the white matter fiber pathways from diffusion MRI data. 
However, fiber-based methods need to preprocess dMRI data to tractography for fiber classification, which is time-consuming and needs large amount of computational sources. Voxel-based tract segmentation methods ~\cite{Wasserthal2018-jj,Lu2021-fp,Lu2022-iv,Liu2022-oa} can directly utilize raw dMRI data to perform tract segmentation. TractSeg is a representative voxel-based white matter tract segmentation method with a U-Net-based structure.  Segmenting tracts with a 3D segmentation model requires huge GPU memory and computing power. TractSeg transforms it into a combined task of multiple 2D slice segmentation. However, tract annotation is time-consuming and laborious. Most few-shot methods for tract segmentation ~\cite{Lu2021-fp,Lu2022-iv,Liu2022-oa} utilize transfer learning or fine-tuning strategies, but one-shot novel tract segmentation is under-studied.  

\subsection{Class Incremental Semantic Segmentation}
Class incremental semantic segmentation (CISS) aims to transfer base knowledge to novel model without old data and has recently attracted attention in the community. LwF \cite{Lwf} designs base and novel heads structure for protecting base knowledge and learning novel knowledge. Building on LwF, PLOP \cite{PLOP} introduces a multi-scale pooling distillation scheme to preserve base knowledge through multi-layer feature distillation. EWF \cite{EWF} further fuse end-point layers of the base model and novel model to better balance base and novel knowledge. However, both PLOP and EWF are designed for single-label segmentation (with the softmax activation function), making them unsuitable for multi-label segmentation tasks, resulting in performance degradation or incompatibility. CoinSeg \cite{CoinSeg} incorporates single-label contrastive learning to improve intra-class consistency and inter-class diversity. However, its single-label contrast strategy leads to conflicts in multi-label segmentation, as illustrated in Figure \ref{fig:1}(b). To the best of our knowledge, no method has been specifically developed for multi-label class incremental segmentation.
\subsection{Voxel-wise Contrastive Learning}
Recently, contrastive learning methods \cite{he2020momentum,he2022masked,chen2021exploring} have outperformed other pretext-task-based methods. The main idea of contrastive learning is to close similar pairs and keep dissimilar pairs away from each other. He et al.~\cite{he2020momentum,he2022masked} define different views of the same image as positive pairs and different images as negative pairs. By designing different loss functions, they narrow positive pairs and push away negative pairs. Wang et al.~\cite{wang2021exploring} introduce the idea of contrastive learning into fully-supervised image segmentation. The main idea is to regard voxels as images: the voxels of the same class are defined as positive pairs and the voxels of different classes are defined as negative pairs. Voxel-wise contrastive learning is carried out through voxel-wise classification loss. 
CoinSeg \cite{CoinSeg} obtains the prototype of each class through Mask Average Pooling (MAP) operation. It adjusts the feature space through single-label voxel contrast loss between the base class and novel classes while protecting the knowledge of base classes.
However, WM tract segmentation is a multi-label segmentation task, which cannot directly distinguish positive/negative voxel pairs. Therefore, it suffers from inherent contradictions.
\section{Methods}

\subsection{Task Definition}
One-shot Class Incremental Semantic Segmentation (OCIS) consists of a series of incremental learning steps, represented as $t = 1,...,T$. Each learning step introduces novel classes using only a single annotated sample, while maintaining knowledge of base classes without forgetting. 
$\{x_{n},y_{n}\}$ are the one-shot tract data and its corresponding multi-label segmentation label. $m$ and $n$ are numbers of base and novel tract classes, respectively.
Set the model of $t$ step as the Novel model: 
\begin{equation}
Y^t= Novel(X_n; \theta^t),
\label{equation:1}
\end{equation} 
where $\theta^t$ is the model parameters of $t$ step.


\subsection{LwF Initialization}
Learning without forgetting (LwF) initialization \cite{Lwf} can transfer the base tract segmentation knowledge and retain the ability of base tract segmentation.
Specifically, let the shared encoders of the base and novel models be $E^{t-1}$ and $E^{t}$, respectively. The base tract segmentation decoder of the base model and the novel tract segmentation decoder of the novel model are denoted as $D^{t-1}_{b}$, $D^{t}_{b}$ and $D^{t}_{n}$, respectively. 
The parameter weights of $E^{t-1}$ and $D^{t-1}_{b}$ are used as the parameter weights of $E^{t}$ and $D^{t}_{b}$, respectively.


\noindent\subsection{Multi-Label Voxel Contrast Framework}
 As shown in Figure~\ref{fig:2}(c), our framework consists of a Base Tract Uncertainty Distillation Module, a Novel Tract Segmentation Module, a Feature Space Voxel Contrast Module, and a Multi-Loss Dynamic Weighting Module. 

\noindent\textbf{Base Tract Uncertainty Distillation Module.}
Although there are some differences between the novel and base tracts, some semantic knowledge learned from the base tracts is class-agnostic. We assume the base model trained on the base tracts could also have semantic knowledge for novel tracts to benefit our one-shot incremental segmentation task on novel tracts. We introduce uncertainty distillation \cite{hao_ISBI23} to improve the efficiency of distillation and enforce the similarity between the predictions of the base tracts from the base and novel models to help the novel model maintain the semantic knowledge from the base model. 
Specifically, first, we freeze the parameters of the base model and obtain an uncertainty map by applying an uncertainty transformation to the base model's predictions for the base tracts:
\begin{equation}
	um=\left\{\begin{matrix}
		2 \cdot  z^{t-1}_{b} -1,& if \quad z^{t-1}_{b}>0.5,\\
		1-  2 \cdot  z^{t-1}_{b},& otherwise.
	\end{matrix}\right.
	\label{equation:8}
\end{equation} where $um$ is the uncertainty map and $z^{t-1}_{b}$ is the output prediction of the base model.
Then, we use this uncertainty map for weighting the distillation loss.

We use a cross-entropy loss as the distillation loss $L_dis$ between the frozen base model and the novel model to measure the base tract segmentation difference between the two models:

\begin{equation}
\begin{split}
L_{dis}&= -\frac{1}{m} \sum_{i=1}^{m} um[i] \{ z^{t-1}_{b}(x_{b})[i] \log_{}{(z^{t}_{b}(x_{b})[i])} \\
 & + (1-z^{t-1}_{b}(x_{b})[i]) \log_{}{(1-z^{t}_{b}(x_{b})[i])}
\},
\end{split}
\label{equation:7}
\end{equation}
where $z^{t}_{b}$ is the base tract prediction of the novel model.

\noindent\textbf{Novel Tract Segmentation Module.}
The binary cross entropy loss is used as the segmentation loss $L_{seg}$ of the novel model for novel tracts.
Set the predictions of the novel model for novel tracts are denoted as $z^{t}_{n}$:
\begin{equation}
\begin{split}
L_{seg}&=-\frac{1}{n} \sum_{j=1}^{n} \{ y_{n}[j] \log_{}{z^{t}_{n}[j]} \\
  & + (1-y_{n}[j]) \log_{}{(1-z^{t}_{n})[j])}
\}.
\end{split}
\label{equation:6}
\end{equation}

\noindent\textbf{Feature Space Voxel Contrast Module.} Novel tract segmentation suffers from a feature overlap problem \cite{song2023learning} in the embedding space. Voxel contrast methods alleviate the above issue by adjusting the feature space through contrast learning. 
Previous works \cite{wang2021exploring,CoinSeg} of voxel-level contrast segmentation are for images on natural scene, which is a single-label voxel-wise segmentation task. It means that they can easily distinguish positive and negative voxel pairs according to whether they belong to the same class, where positive samples are pulled closer and negative samples are pushed apart. However, when samples exhibit semantic overlap, this approach can lead to conflicts. A sample may have multiple labels with shared semantic features, causing incorrect simultaneous pulling and pushing of the same anchor, which disrupts the optimization. To address this, we propose a voxel-wise multi-label contrastive loss which introduces an inter-class similarity measure to more effectively handle semantically overlapping samples. Our approach moves beyond binary positive-negative distinctions by adjusting sample proximity based on label similarity, preventing overly extreme pushing or pulling, and resulting in well-organized feature embedding space so as to alleviate the feature overlap problem. Moreover, it is time-consuming to calculate voxel-wise multi-label contrastive loss directly for each batch. Therefore, we propose an accelerating calculation of the loss to reduce the training time.

Specifically, first, we random sample some voxel-wise embeddings, one of which is taken as the anchor sample. We use the label similarity between this anchor sample and other samples to control the embedding distance between the anchor sample and other samples. We identify positive/negative pairs by the normalized voxel-label similarity and normalized voxel-level feature distance in the embedding space. A voxel pair is defined as the negative pair if its feature distance is too close relative to the voxel label similarity (i.e., inside the orbit, as shown in Figure.\ref{fig:1}(c), and as the positive pair if the feature distance is too far relative to the voxel label similarity (i.e., outside the orbit, as shown in Figure.\ref{fig:1}(c). 
Let a voxel embedding be $z$ and its corresponding voxel-level multi-label be $\hat{y}$. Set $C_{pq} =  {\hat{y}_p}^T \cdot \hat{y}_q$   is the label similarity of a voxel-pair and $g(p)=\{k|k\in{1,2,...,m+n},k\ne p \} $ is a set of voxels except for the voxel $p$. The voxel-wise multi-label contrastive loss $L_{CL}$ is as follow:
\begin{equation}
\begin{split}
{L_{VC}}^{pq} = -\beta_{pq} \log_{}{ \frac{e^{-d(z_p,z_q)/\tau }} {{\textstyle \sum_{k\in g(p)}^{}}  e^{-d(z_p,z_k)/\tau }} },  
\end{split}
\label{equation:8}
\end{equation}
where $\beta_{pq}$ is the dynamic coefficient, $d(\cdot,\cdot)$ is the Euclidean distance, and $\tau$ is a hyperparameter that is set to 1. 

\noindent\textbf{Dynamic Coefficient Adjustment.}
The dynamic coefficient \(\beta\) adjusts the weight of the current voxel pair in the contrastive loss \(L_{VC}\). Specifically, \(\beta_{pq}\) is defined as the ratio of the label similarity between voxel pair \(p\)-\(q\) to the sum of label similarities of all other voxel pairs in the current batch:
\begin{equation}
\begin{split}
{\beta_{pq} = \frac{C_{pq}}{\sum_{k\in g(p)} C_{pk}}}.
\end{split}
\label{equation:9}
\end{equation}
If the label similarity of voxel pair \(p\)-\(q\) is higher, \(\beta_{pq}\) becomes larger, encouraging the model to pull these voxels closer in the feature embedding space, and vice versa. This normalization prevents the model from excessively pulling or pushing specific voxels, thereby ensuring a balanced distribution in the feature space.

\noindent\textbf{Voxel-wise Sampling Strategy and its Acceleration.}
Previous research \cite{robinson2020contrastive,hu2021adco,zhou2021c3} found that the quality of samples is critical to voxel contrast loss and the region around the boundary of the mask is difficult to segment. Depending on whether the samples are near the center or boundary of each tract mask, each image is divided into three regions. One is the region in the mask far from the boundary beyond the threshold, which is called the inner area. Another is the region on both sides of the boundary that is less than the threshold, called the outer region. The other is the region far from the tract mask boundary beyond the threshold, which is called the background area. For all tract masks of one image, we randomly take an anchor sample in the inner region of each tract mask. For each anchor, we randomly take the same number $n_s$ of inner samples, outer samples, and background samples. 

In addition, we find that it is very time-consuming to calculate the voxel-wise contrastive loss for all tract classes and the whole batch. To accelerate the training time, we adopt two tricks. First, if a slice does not contain a certain tract class (the tract mask is all negative), we do not take the anchor and its corresponding samples. Second, computing the voxel contrast loss for the entire batch is not necessary. Therefore,  we only sample from a part of the batch to reduce the calculation cost. Specifically, we first calculate the sum of $L_{dis}$ and $L_{seg}$ of each batch and sort it. Then, we averagely sample one-eighth of the batch by size according to the sorted batch to speed up the training time.

\noindent\textbf{Multi-Loss Dynamic Weighting Module.}
The overall loss of our method is composed of multiple losses. Empirically, our performance is highly dependent on the relative weighting of each loss. Manually tuning these weights is both challenging and time-consuming. To address this, we introduce a dynamic weighting module \cite{dynamic_weighting}\footnote{See \cite{dynamic_weighting} for details of dynamic weighting}, which balances the multiple loss functions by accounting for the homoscedastic uncertainty of each task. This approach enables simultaneous learning across multiple losses, even when they have different units or scales, improving the efficiency and effectiveness of the model. Through the dynamic weighting module, we obtain the weights of $L_{seg}$, $L_{dis}$, and $L_{VC}$ are $w_1$, $w_2$, $w_3$, respectively.

\begin{table*}[th]
	\centering
    
	\fontsize{8.5}{9.5}\selectfont
		\caption{Comparison results of our proposed method against other state-of-the-art methods on the HCP dataset using DSC (\%). The best results are highlighted in bold. More results on Preto datasets can be found in the Appendix.}
		\label{tabel:1}
        \vspace{-0.3cm}  
		\resizebox{\linewidth}{!}{ 
			\begin{tabular}{l| ccc |ccc| ccc| ccc |ccc} 
				\toprule[0.5mm]
				& \multicolumn{3}{c|}{\textbf{60-12 (2 steps)}} & \multicolumn{3}{c|}{\textbf{48-24 (2 steps)}} & \multicolumn{3}{c|}{\textbf{36-36 (2 steps)}} & \multicolumn{3}{c|}{\textbf{24-48 (2 steps)}} & \multicolumn{3}{c}{\textbf{12-60 (2 steps)}} \\ 
              
				\multirow{-2}{*}{\textbf{Method}} & 1-60 & 61-72 & all & 1-48 & 49-72 & all & 1-36 & 37-72 & all & 1-24 & 25-72 & all & 1-12 & 13-72 & all \\
				\midrule[0.3mm]
				VM-DA \cite{Balakrishnan2019-zj} & 59.3 & 59.4 & 59.3 & 59.3 & 59.2 & 59.3 & 59.9 & 59.5 & 59.8 & 59.1 & 57.8 & 58.2 & 59.0 & 59.2 & 59.1 \\
				CFT \cite{Lu2022-iv} & 70.9 & 63.9 & 69.8 & 70.1 & 54.7 & 65.0 & 66.6 & 58.2 & 62.4 & 71.8 & 50.4 & 57.5 & 72.4 & 51.3 & 54.8 \\
				IFT \cite{Lu2022-iv} & 69.1 & 77.2 & 70.4 & 69.2 & 67.4 & 68.6 & 67.5 & 59.6 & 63.5 & 70.2 & 24.1 & 39.5 & 72.5 & 14.8 & 24.4 \\
				TractSeg-LE \cite{Liu2022-oa} & 67.4 & 48.5 & 64.2 & 62.1 & 48.7 & 57.6 & 60.8 & 57.7 & 62.7 & 61.6 & 47.8 & 52.4 & 68.7 & 44.4 & 48.5 \\
				LwF \cite{Lwf} & 75.0 & 66.9 & 73.6 & 73.9 & 67.6 & 71.8 & 69.9 & 66.8 & 68.4 & 72.7 & 63.9 & 66.8 & 70.5 & 60.6 & 62.2 \\
				PLOP \cite{PLOP} & 72.7 & 65.7 & 71.6 & 71.4 & 63.7 & 68.8 & 64.7 & 61.7 & 63.2 & 67.9 & 56.5 & 60.3 & 65.8 & 56.0 & 57.6 \\
				EWF \cite{EWF} & 57.6 & 60.7 & 58.2 & 55.5 & 58.4 & 56.5 & 51.1 & 54.3 & 50.7 & 53.5 & 50.8 & 51.7 & 52.9 & 50.5 & 50.9 \\
                    MCLOS \cite{MCLOS} &76.9&68.2 &72.5&74.3& 64.9 &69.6 &74.0 &61.3 &67.7 & 76.2&59.4 &67.8& 74.6& 55.8& 62.2\\
				CoinSeg \cite{CoinSeg} & 79.8 & 75.5 & 79.1 & 75.4 & 70.4 & 73.7 & 73.7 & 65.1 & 69.4 & 80.1 & 60.5 & 67.0 & 79.4 & 58.5 & 62.0 \\
				\midrule[0.2mm]
                \rowcolor{mygray}
				\textbf{MultiCo3D (Ours)} & \textbf{81.6} & \textbf{82.0} & \textbf{81.7} & \textbf{81.5} & \textbf{75.5} & \textbf{79.5} & \textbf{80.1} & \textbf{75.3} & \textbf{77.7} & \textbf{82.8} & \textbf{71.4} & \textbf{75.2} & \textbf{84.8} & \textbf{72.3} & \textbf{74.4} \\
				\bottomrule[0.5mm]
			\end{tabular}
		}
  \vspace{-0.2cm}
\end{table*}

\begin{table*}[]
\caption{Ablation study on our LwF initialization (\textit{LwF}), distillation loss (\textit{$L_{dis}$}), voxel-wise multi-label contrastive loss (\textit{$L_{VC}$}) and Dynamic Weighting (\textit{DW}). The best results are highlighted in bold.}
\centering
\vspace{-0.3cm}  
\fontsize{8.5}{9.5}\selectfont
\resizebox{\linewidth}{!}
{
\begin{tabular}{ cccc |ccc| ccc| ccc| ccc| ccc }
 \toprule[0.5mm]
& & & 
 & \multicolumn{3}{c|}{\textbf{60-12 (2 steps)}}                         & \multicolumn{3}{c|}{\textbf{48-24 (2 steps)}}                         & \multicolumn{3}{c|}{\textbf{36-36 (2 steps)}}                   & \multicolumn{3}{c|}{\textbf{24-48 (2 steps)}}                    & \multicolumn{3}{c}{\textbf{12-60 (2 steps)}}   \\ 
              
\multirow{-2}{*}{\textit{LwF } }&\multirow{-2}{*}{\textit{$L_{dis}$ }} &\multirow{-2}{*}{ \textit{$L_{VC}$ }} & \multirow{-2}{*}{\textit{DW} } &

1-60       & 61-72          & all              & 1-48            & 49-72    & all      & 1-36  &37-72  & all   & 1-24    & 25-72              & all            & 1-12& 13-72                 & all               \\  
\midrule[0.3mm]
                                                                       
 \cmark & \xmark & \xmark  &\xmark  &  74.97          & 66.91          & 73.63          & 73.88           & 67.63         & 71.80         & 69.91           & 66.79 & 68.35 & 72.70           & 63.87   & 66.81  & 70.46 & 60.58 & 62.23\\ 

 \cmark & \cmark & \xmark  &\xmark & 80.72 & 77.68 & 80.21 & 80.32 & 74.13 & 78.26 & 77.17 & 72.33  & 74.75  & 80.65 & 69.73 & 73.37 & 81.15 & 69.26 & 71.24   \\ 

\cmark & \cmark & \cmark  &\xmark& 81.45 & 78.24 & 80.92 & 80.33 & 75.34 & 78.67 & 77.71 & 74.98   & 76.34 & 80.77 & 71.03 & 74.28 & 81.56       & 70.54 & 72.38 \\

\rowcolor{mygray}
 \cmark & \cmark & \cmark  &\cmark & \textbf{81.62} & \textbf{81.95} & \textbf{81.68} & \textbf{81.53} & \textbf{75.52} & \textbf{79.53} & \textbf{80.12} & \textbf{75.26} & \textbf{77.69} & \textbf{82.77} & \textbf{71.37} & \textbf{75.17} & \textbf{84.78} & \textbf{72.31} & \textbf{74.39} \\ 
 \bottomrule[0.5mm]
\end{tabular}
\label{tabel:2}
}
\vspace{-0.4cm}
\end{table*}

\begin{figure*}[ht]
  \centering
   \includegraphics[width=0.98\linewidth]{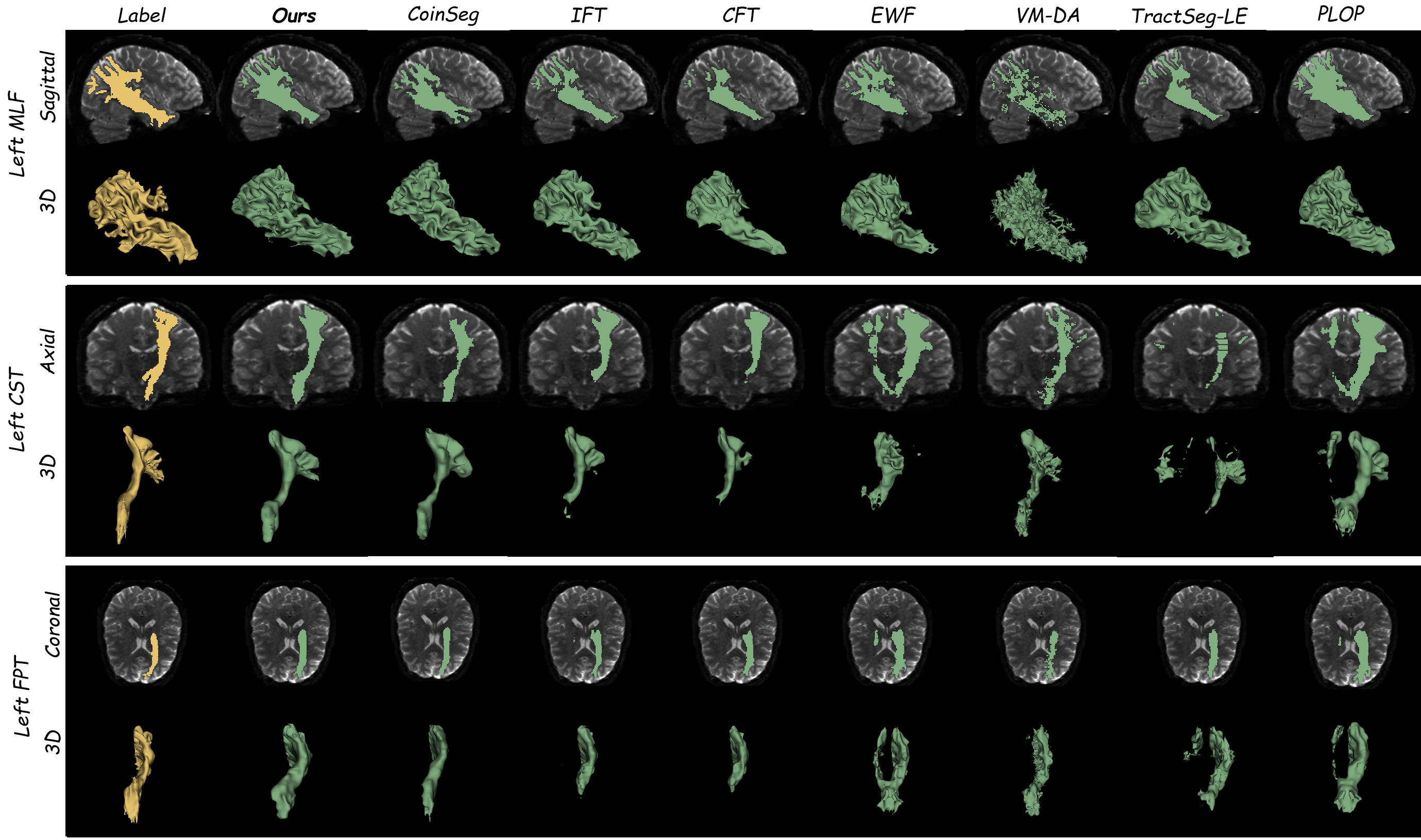}
   \vspace{-0.2cm}  
   \caption{Visualization of segmentation results of three novel tracts: left middle longitudinal fascicle (MLF), left corticospinal tract (CST), and left frontopontine tract (FPT) on 36-36 (2 steps) setting. The yellow regions are labels, and the green regions are segmentation results of our method and compared methods.}
   \label{fig:3}
   \vspace{-0.5cm}
\end{figure*}

To sum up, we use the sum of three losses as the total loss: 
\begin{equation}
\begin{split}
L =w_1 L_{seg} +w_2 L_{dis}+w_3 L_{VC}.  
\end{split}
\label{equation:9}
\end{equation}
The distillation loss protects the base tract knowledge while the model learns to segment novel tracts. The multi-label voxel contrast loss, by introducing label similarity, adjusts the feature space more precisely, avoiding the contradictions of previous single-label methods and mitigating the issue of feature overlap. Finally, the dynamic weighting module balances different losses more efficiently, driving the model to learn more effectively and enhancing its overall performance.

\section{Experiments and Results}

\subsection{WM Tract Dataset and Data Preprocessing}
\noindent\textbf{HCP Dataset.} The HCP dataset is from \cite{Wasserthal2018-jj}, which contains 105 subjects from the Human Connectome Project (HCP) \cite{Van_Essen2013-sy}. Each subject has 72 tracts annotated by neuroanatomists. We random split 72 tracts as base and novel tracts based on different brain regions. Specifically, we first group the 72 tracts by brain region into six sets, each containing 12 tracts. Then, based on the experimental setup, we randomly split these sets to either base or novel tracts. 
For the two-step incremental setup, 63 subjects are used for training base tracts, only one subject for training novel tracts, and 20 and 21 subjects for validation and testing, respectively. The full name of each tract can be seen in the Appendix. In addition, we also conduct our experiments on the Preto \cite{Preto} dataset. 
The Preto \cite{Preto} dataset is a diffusion MRI dataset with a large number of samples for 20 healthy individuals. All the subjects were acquired at the Hospital das Clínicas at Ribeirão Preto in Brazil on a 3.0T MRI scanner. We use MITK \cite{MITK} to generate 72 tracts for our segmentation task. For the two-step incremental setup, we split 20 subjects into 9 for base tract training, 1 for novel tract training, 5 for validation, and 5 for testing. 
More experimental results on Preto dataset can be found in the Appendix.

\noindent\textbf{Data Preprocessing.} We use the multi-shell multi-tissue constrained spherical deconvolution (CSD) method \cite{Tournier2007-cn} with all gradient directions for transforming dMRI data to fiber orientation distribution function (fODF) peak data, which has 9 channels corresponding to sagittal, axial, and coronal directions (each direction has three channels). The fODF peak data is used as the input of tract segmentation model, which achieves higher accuracy compared with raw dMRI data \cite{Wasserthal2018-jj}.

\subsection{Implementation Details}

We use TractSeg \cite{Wasserthal2018-jj} as our base segmentation model due to its excellent performance in white matter tract segmentation. The input to the model is the 2D slices of fODF peak data, with the fODF data dimensions being 144 × 144 × 144 for the HCP dataset and 128 × 128 × 128 for the Preto dataset. 
The nominal resolution is 1.25 mm isotropic for HCP dataset and 2.0 mm isotropic for Preto dataset. 
Each 2D slice includes 9 channels, corresponding to three dimensions in three directions (sagittal, axial, and coronal). During training, the Adamax optimizer is employed, with the number of epochs set to 200 for the HCP dataset and 500 for the Preto dataset. The learning rate is set to 0.002, the dropout rate to 0.4, and the hyperparameters $\tau$ and $n_s$ are set to 1 and 5, respectively. The batch sizes for the HCP and Preto datasets are 48 and 72, respectively.

In the testing phase, each subject is divided into 144 2D slices for the HCP dataset and 128 slices for the Preto dataset in the sagittal, axial, and coronal directions, which are then fed into the model. Predictions in each direction are stacked to create a 3D prediction map for the whole brain. The final white matter tract segmentation is derived by averaging the predictions across the three directions. Our experiment is performed with Pytorch \cite{Pytorch} (v1.10) on an NVIDIA GeForce RTX 4090 GPU machine.

\subsection{Quantitative and Qualitative Results}
We perform comparison experiments and ablation study on the testing set. Our performance evaluation is based on the widely used Dice Score Coefficient (DSC) metric ~\cite{Wasserthal2018-jj,Li2020-nt,Lu2021-fp,Lu2022-iv}. We evaluate our methods on five incremental scenarios: 60-12 (2 steps), 48-24 (2 steps), 36-36 (2 steps), 24-48 (2 steps), and 12-60 (2 steps).
We compare our method with VM-DA \cite{zhao2019data}, classic fine-tuning method (CFT) \cite{Lu2022-iv}, improved fine-tuning method (IFT) \cite{Lu2022-iv}, TractSeg with label embedding (TractSeg-LE) \cite{Liu2022-oa}, Learning without Forgetting (LwF) \cite{Lwf}
Pseudo labeling and LOcal Pod (PLOP) \cite{PLOP},
Endpoints Weight Fusion (EWF) \cite{EWF},
and CoinSeg \cite{CoinSeg} in these five settings on the HCP and Preto datasets. 
We select TractSeg~\cite{Wasserthal2018-jj}) as our step 1 model for base tract segmentation. 
VM-DA, CFT, IFT, and TractSeg-Le are designed for one-shot tract segmentation. We reimplement them by adding LwF scheme for incremental one-shot tract segmentation.
LwF, PLOP, EWF, and CoinSeg are designed for single label one-shot semantic segmentation. We reimplement them by changing their softmax action layer into the sigmoid layer.



\afterpage{
\begin{figure*}[tp]
    \centering
    \begin{subfigure}[b]{0.24\textwidth}
        \centering
        \includegraphics[width=0.98\linewidth]{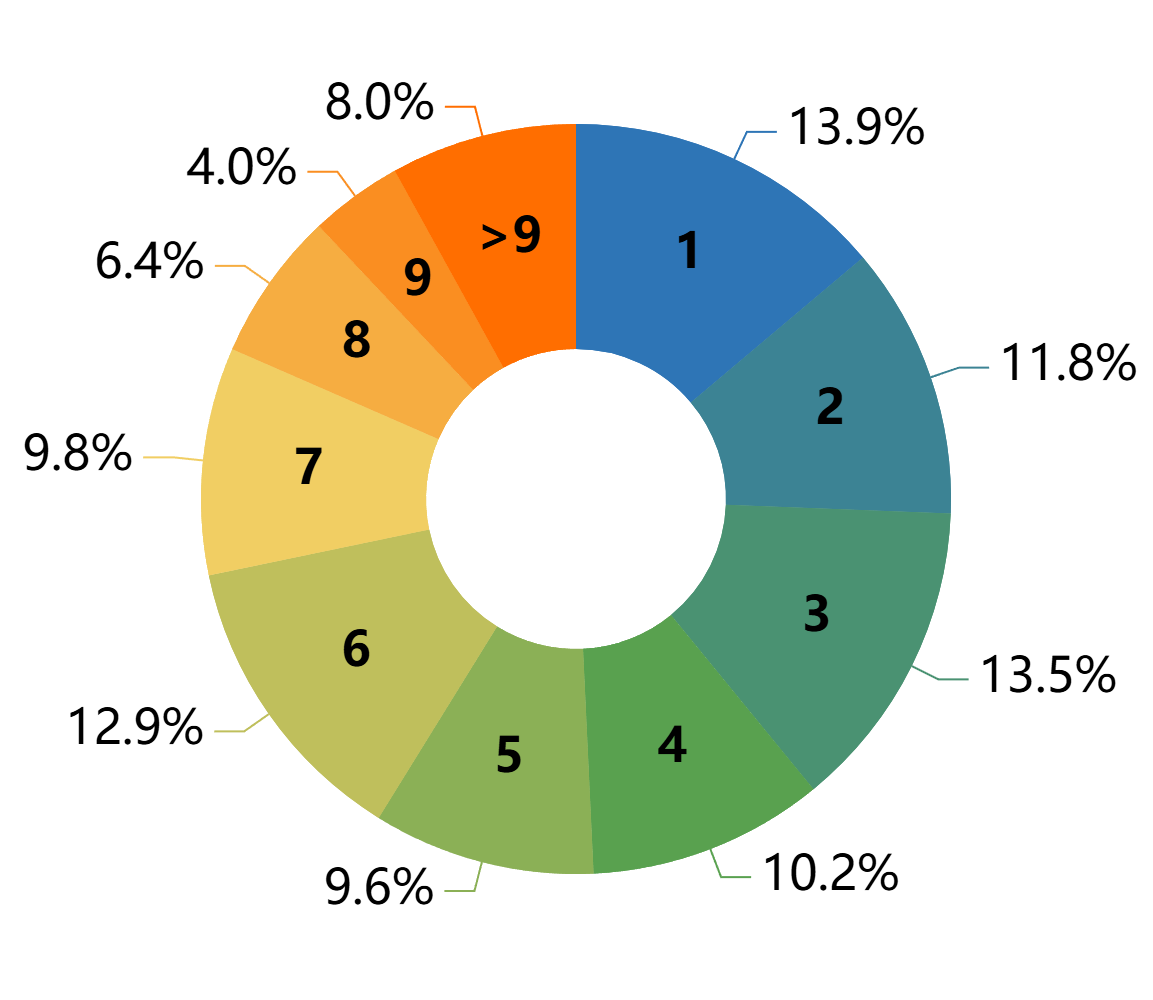} 
          \caption{}
        \label{fig:sub1}
    \end{subfigure}
    \hfill
    \begin{subfigure}[b]{0.24\textwidth}
        \centering
        \includegraphics[width=0.98\linewidth]{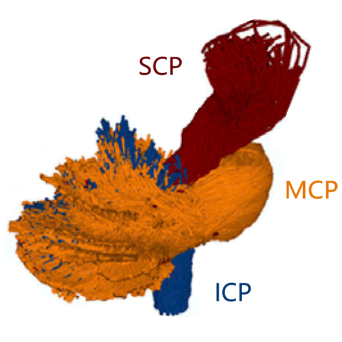} 
          \caption{}
        \label{fig:sub1}
    \end{subfigure}
    \hfill
    \begin{subfigure}[b]{0.24\textwidth}
        \centering
        \includegraphics[width=0.98\linewidth]{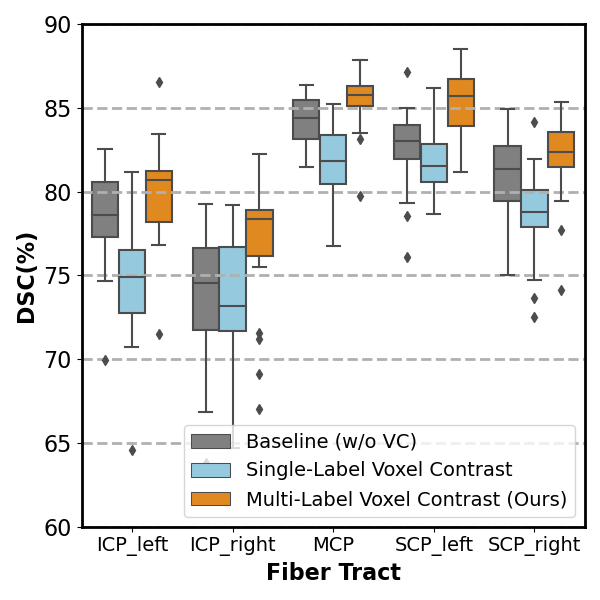} 
          \caption{}
        \label{fig:sub2}
    \end{subfigure}
    \hfill
    \begin{subfigure}[b]{0.24\textwidth}
        \centering
        \includegraphics[width=0.98\linewidth]{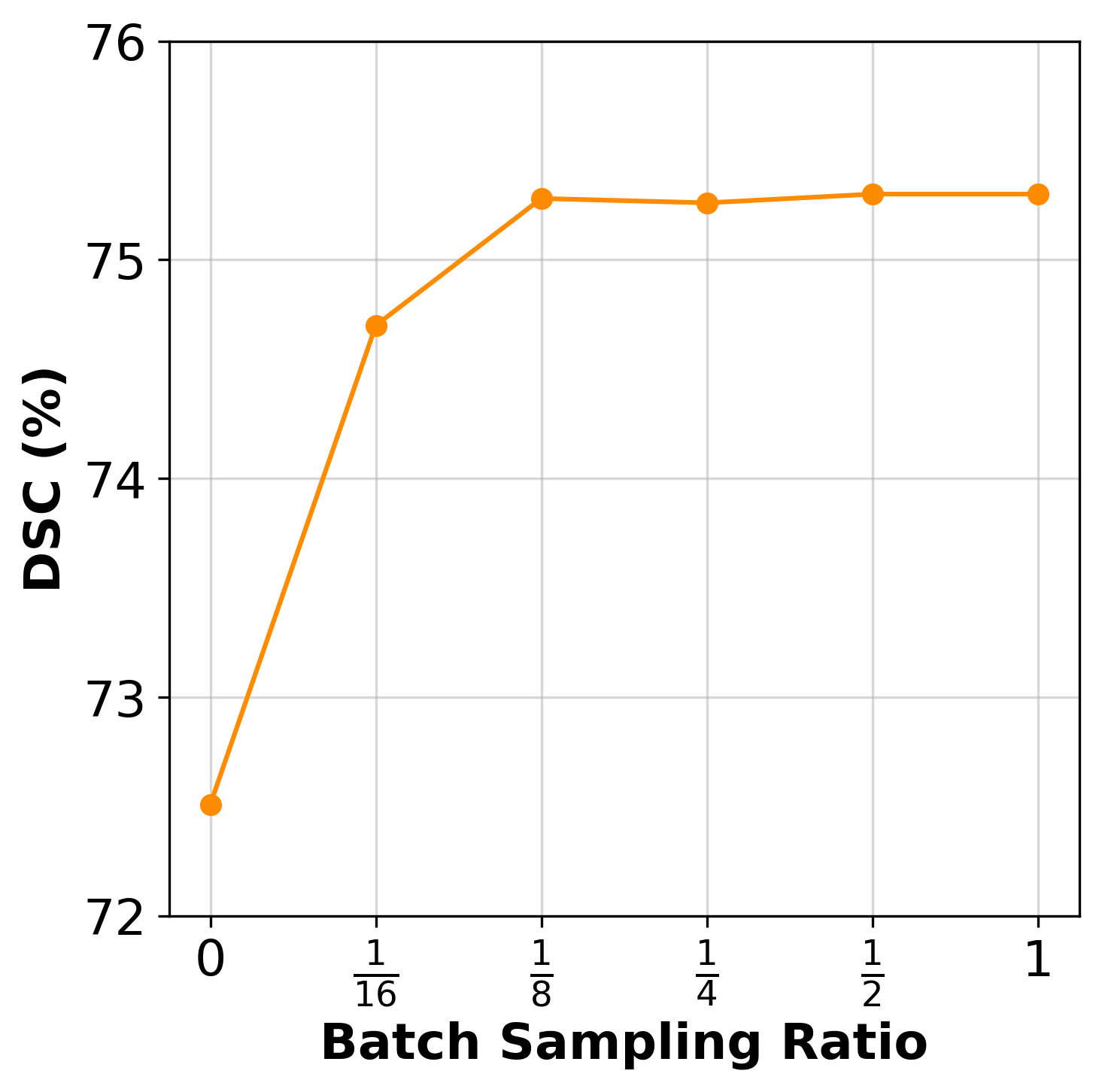} 
        \caption{}
        \label{fig:sub2}
    \end{subfigure}
    \vspace{-0.3cm}
    \caption{Further analysis of our proposed method. (a) The ratio of multi-label voxel overlap of the HCP dataset. (b) Visualization of the streamline overlap of a set of tracts. (c) Performance comparison between different sampling strategies on this set of tracts. (d) Performance comparison between different batch sampling ratio in $L_{VC}$ under the 36-36 (2 steps) setting.}
    \vspace{-0.5cm}
    \label{fig:4}
    
\end{figure*}

}

\noindent\textbf{Comparison with SOTA Methods on the HCP Dataset.}
As shown in Table~\ref{tabel:1}, \textbf{\textit{MultiCo3D (ours)}} significantly outperforms other SOTA approaches by a huge margin in all incremental scenarios. For example, \textbf{\textit{MultiCo3D (ours)}} outperforms LwF, PLOP, EWF, and CoinSeg up to 9.34\%, 14.48\%, 27.01\%, and 8.3\% DSC, respectively. In addition, PLOP and EWF gradually decline, because their design is not suitable for multi-label segmentation tasks. Our method is specifically designed for the multi-label segmentation tasks and thus achieves good results.

\noindent\textbf{Analysis of Multi-Label Voxel Overlap.}
We report the ratio of multi-label voxel overlap of the HCP dataset, as shown in Figure \ref{fig:4}(a). The categories represent the number of labels per voxel, excluding background (unlabeled) voxels. Over 90\% of voxels exhibit multi-label overlap, with more than 50\% of voxels showing high overlap ( $\ge $ 5 labels). This result demonstrates that fiber tract overlap is highly prevalent in neuroimaging data, making single-voxel contrastive methods insufficient for effectively handling such complex multi-label overlap scenarios.

\subsection{Ablation Studies}

\noindent\textbf{Overall Ablation of MultiSeg3D.}
We investigate the effectiveness of our proposed \textit{LwF}, \textit{$L_{dis}$}, \textit{$L_{VC}$}, and \textit{DW} in all five settings. As shown in Table~\ref{tabel:2}, using only \textit{LwF} achieves the promising segmentation performance of novel tracts. Using \textit{LwF + $L_{dis}$} together can significantly improve the performance on base and novel tracts. It demonstrates that the combined utilization of \textit{LwF} and \textit{$L_{dis}$} can not only distillate the base tract segmentation knowledge from the base model to the novel model but also transfer tract segmentation semantic knowledge from base tracts to novel tracts. Compared with \textit{LwF + $L_{dis}$}, \textit{LwF + $L_{dis}$ + $L_{VC}$} can further increase the DSC of novel tracts in all five setups by up to 2.65\%, which proves that \textit{$L_{VC}$} can adjust the embedding space to improve the segmentation performance of novel tracts without degrading the segmentation ability of base tracts. Finally, \textit{LwF + $L_{dis}$ + $L_{VC}$ + DW} further improve the DSC of all five setups by up to 3.71\%, which proves the effectiveness of the dynamic weighting module.

\begin{table}[h]
\centering

\caption{Performance comparison of different sampling strategies on DSC (\%) under the 36-36 (2 steps) setting. The best results are highlighted in bold.}

\resizebox{\linewidth}{!}{

\begin{tabular}{l|ccc}
\toprule[0.5mm]
\textbf{Sampling Strategy} & \textbf{1-36} & \textbf{37-72} & \textbf{all} \\ 

\midrule[0.3mm]
Baseline (\textit{w/o} Voxel Contrast) & 79.0 & 72.5 & 75.8 \\ \hline
Random & 73.8$_{\textcolor{red}{-5.2}}$ & 62.2$_{\textcolor{red}{-10.3}}$ & 68.0 $_{\textcolor{red}{-7.8}}$ \\
Balanced & 79.2$_{\textcolor{mygreen}{+0.2}}$ & 73.4$_{\textcolor{mygreen}{+0.9}}$ & 76.3$_{\textcolor{mygreen}{+0.5}}$ \\
\rowcolor{mygray} 
\textbf{Balanced + Hard (Ours) }& \textbf{80.1}$_{\textcolor{mygreen}{+1.1}}$ & \textbf{75.3}$_{\textcolor{mygreen}{+2.8}}$ & \textbf{77.9}$_{\textcolor{mygreen}{+2.1}}$ \\
 \bottomrule[0.5mm]

\end{tabular}
}
\vspace{-0.5cm}
\label{tabel:3}
\end{table}

\noindent\textbf{Ablation Study of Different Sampling Strategies.}
Table~\ref{tabel:3} shows the effect of different sampling strategies.
\textbf{\textit{Random}} represents random sampling without distinguishing between positive and negative samples. \textit{\textbf{Balanced}} refers to sampling the same number of positive and negative samples.
\textbf{\textit{Hard}} is sampling the same number of simple and hard samples according to the segmentation label. We observe that \textit{\textbf{Balanced + Hard (Ours)}} achieves the highest DSC compared with \textit{\textbf{Random}} and \textit{\textbf{Balanced}}, evaluating the effectiveness of our sampling strategy.

\noindent\textbf{Ablation Study of Batch Sampling Ratio of $L_{VC}$.}
As shown in Figure \ref{fig:4}(d), we show the results of different batch sampling ratios. Batch sampling ratio means that the proportion selected from a whole batch is used to calculate $L_{VC}$.
We observe that our method achieves the highest DSC with the smallest batch sampling ratio of one-eighth. This means that only one eighth of the batch is needed to achieve the same effect as the entire batch, which speeds up the calculation.

\noindent\textbf{Quantitative Comparison of Single-Label and Multi-Label Voxel Contrast.} To analyze which tracts are benefit from Multi-Label Voxel Contrast Strategy (Ours), we visualize the ICP, MCP, and SCP tracts which are highly overlapping in Figure~\ref{fig:4}(b), and show the quantitative comparison of Single-Label Voxel Contrast and Multi-Label Voxel Contrast in Figure~\ref{fig:4}(c). The result shows that Single-Label strategy has a negative effect on ICP, MCP, and SCP tracts, while our Multi-Label Voxel Contrast Strategy improves segmentation performances of these highly overlapping tracts. More experimental results for other tracts can be found in the Appendix. 


\subsection{Visualization of Tract Segmentation}
Figure~\ref{fig:3} shows the segmentation results of three novel tracts in axial, coronal, and sagittal three planes and the corresponding 3D view in the 36-36 setup.  We observe that the shape of VM-DA's segmentation results is close to the ground truth (GT), but its 3D surface is not smooth and has lots of noise on it. The possible reason is that VM-DA are based on registration, therefore cannot accurately segment specific tracts for each subject due to inter-subject variability. The 3D surface of CFT and IFT is relatively smooth but incomplete. Their CST-left results are missing the bottom part compared to the GT. 
The shape of TractSeg-LE, EWF, and PLOP is quite different from the GT. CoinSeg lacks some detail textures in 3D perspective. In contrast, the segmentation result of our method not only has a visually smooth 3D surface but also has less over-segmentation and noise.

  

\section{Conclusion}
In this work, we propose a novel multi-label voxel-contrast-based framework for one-shot incremental tract segmentation. Our framework proposes an uncertainty-based distillation module to protect the base tract knowledge while learning novel tract segmentation knowledge, a multi-label voxel contrast module to adjust the feature space and alleviate the feature overlap problem, and a dynamic weighting module to balance the multi-task losses for improving the model performance. Comparison and ablation experiments in multiple experimental settings demonstrate the effectiveness of our proposed method. Our results illustrate the potential utility of one-shot incremental learning framework for WM tract segmentation. Our framework can be particularly valuable for studies with scarce tract labels (e.g., superficial white matter) or studies where high-quality annotations are difficult and expensive to obtain (e.g., high-resolution dMRI data). In future work, we will implement our framework on few-shot settings to further boost performance. 



\section*{Appendix}

\section*{A. Data Augmentation}
The data augmentation methods used by the SOTA methods and our method following the previous work~\cite{Wasserthal2018-jj} as below:

\noindent$\bullet$ Rotation by angle $\Phi \sim U[-\frac{\pi}{4},\frac{\pi}{4}]$. 

\noindent$\bullet$ Zooming by a factor $\lambda \sim U[0.9, 1.5]$. 

\noindent$\bullet$ Random flipping.

\noindent$\bullet$ Displacement by $\triangle x \sim U[-10, 10]$, $\triangle y \sim U[-10, 10]$.

\noindent$\bullet$ Gaussian noise with mean and variance $(\mu, \delta) \sim (0, U[0, 0.05])$.

\section*{B. Inference Time} 

We show the inference time for each test subject of our method and the comparison method in the 36-36 (2 steps) setting on the HCP datasets, as shown in Table ~\ref{tabel:s1}. Since our method is relatively complex, the inference time of our method lags behind the SOTA methods, but is within an acceptable range.

\section*{C. Quantitative Comparison of Different Voxel Contrast Strategy on the HCP Dataset}
\begin{figure*}[bp]
  \centering
   \includegraphics[width=0.99\linewidth]{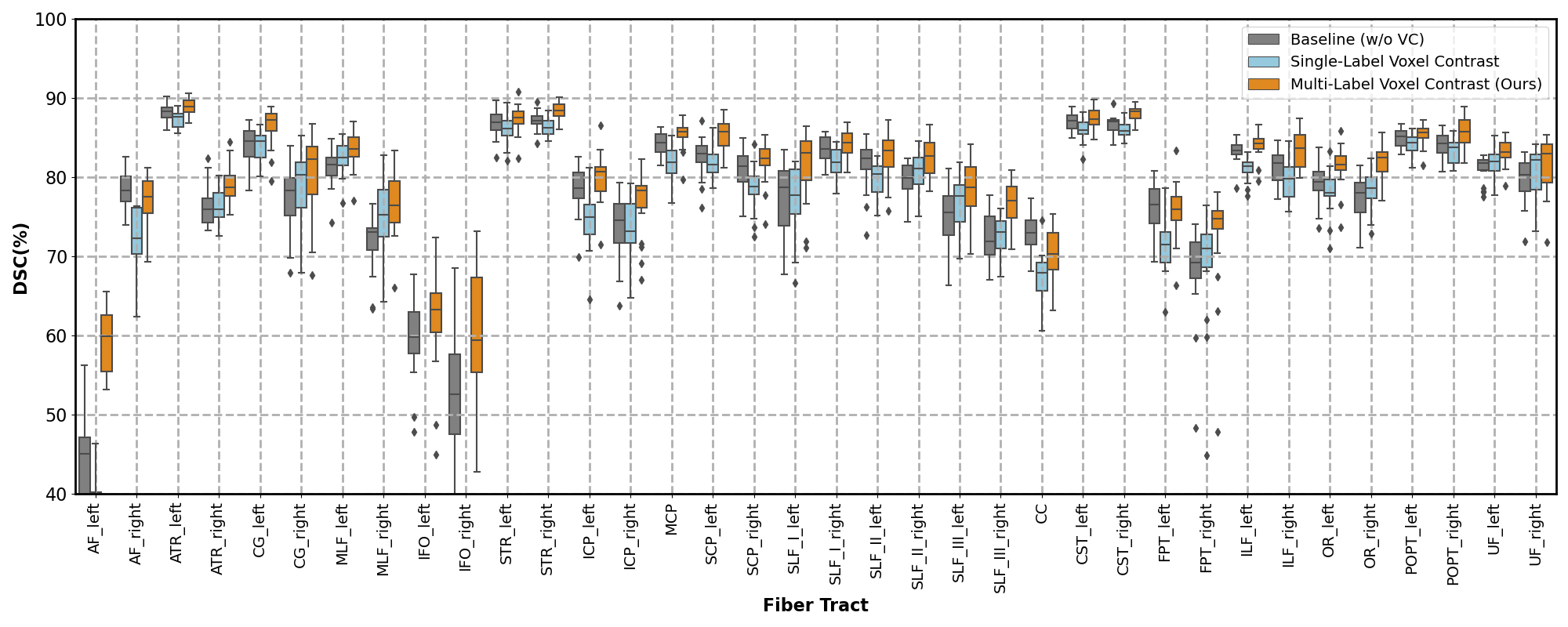}
   \caption{Quantitative comparison of Single-Label Voxel Contrast and Multi-Label Voxel Contrast of 36 tracts using DSC (\%) under the 36-36 (2 steps) setting on the HCP dataset.}
   \label{fig:s1}
\end{figure*}

As shown in Figure \ref{fig:s1}, the boxplot indicates the DSC of \textit{baseline method (w/o VC)}, \textbf{Single-Label Voxel Contrast}, and \textbf{Multi-Label Voxel Contrast} over 36 novel tracts. We observe that \textbf{Single-Label Voxel Contrast} has a negative impact on our multi-label segmentation task, while our \textbf{Multi-Label Voxel Contrast} improves the segmentation performance of novel tracts, which demonstrates the superiority of our \textbf{Multi-Label Voxel Contrast} strategy.

\section*{\vspace{3mm}
\\ 
D. Comparison with SOTA Methods of Five Settings on the Preto Dataset}

\begin{figure*}[htbp]
  \centering
   \includegraphics[width=0.99\linewidth]{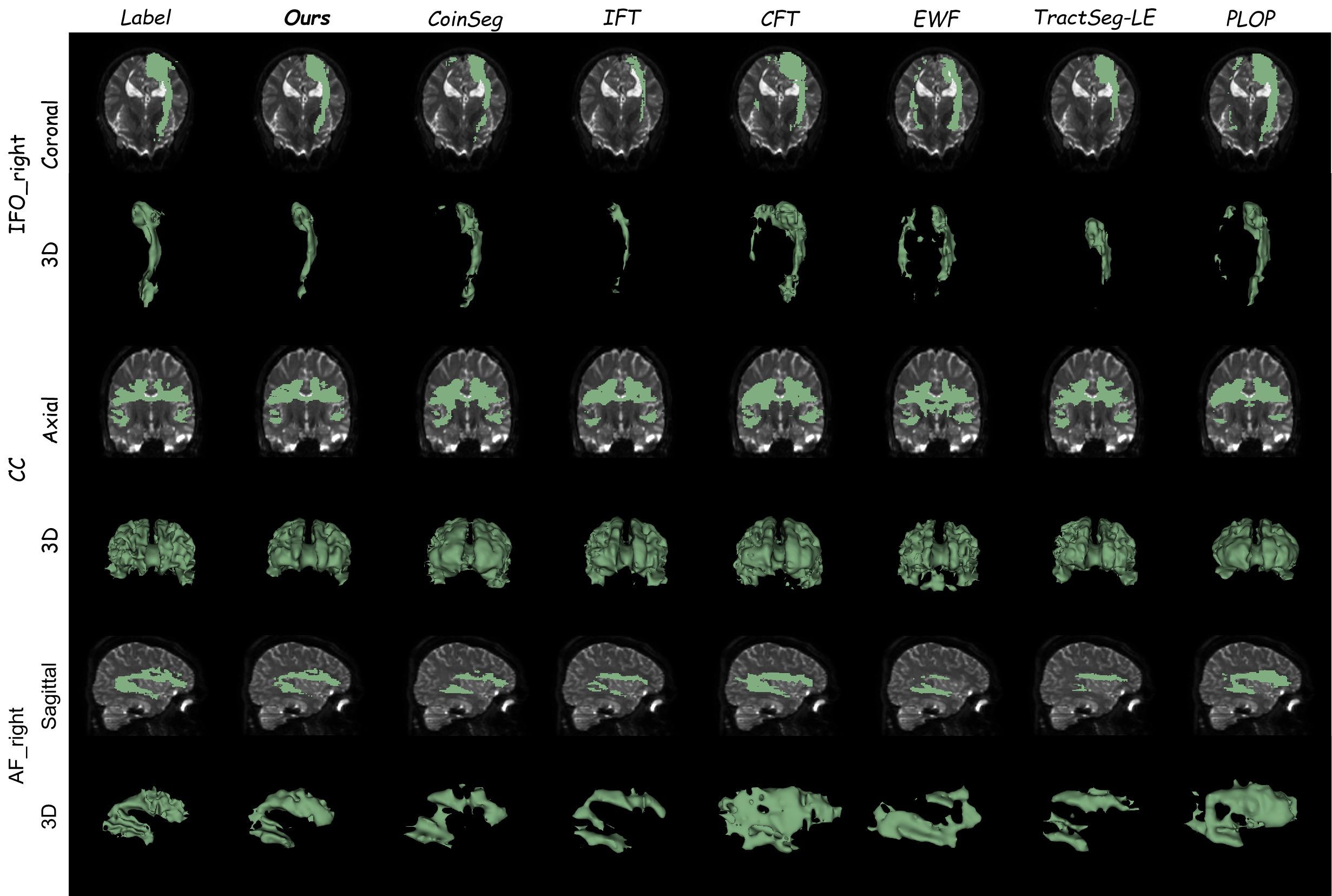}
   \caption{Visualization of segmentation results of three novel tracts: Right Inferior Occipito-Frontal Fascicle (IFO right), Corpus Callosum (CC), and Right Arcuate Fascicle (AF right) on 36-36 (2 steps) setting on the Preto dataset. }
   \label{fig:s2}
\end{figure*}

We compare our method with the SOTA methods in five settings on the Preto dataset. As shown in Table ~\ref{tabel:s2}, our method surpasses the SOTA methods in all five settings for base and noxel tract classes. Specifically, our method exceeds CoinSeg up to 7.12\% DSC. Moreover, we visualize the segmentation results of our method and the SOTA methods on the Preto dataset, as shown in Figure \ref{fig:s2}. Our method is most similar to the label, with complete shapes and relatively smooth but rich textures. 
In addition, Table \ref{tab:S3} shows the abbreviations and full names of the 72 tracts used in the HCP and Preto datasets.

\begin{table}[htbp]

\caption{Quantitative comparison of inference time of SOTA methods under the 36-36 (2 steps) setting on the HCP dataset. The best results are highlighted in bold.}
\centering
\resizebox{0.6\linewidth}{!}
{
\begin{tabular}{l|c}

\toprule[0.5mm]
\textbf{Method}    & \textbf{Time (s)} \\ \midrule[0.3mm]

CFT~\cite{Lu2022-iv}        
& \textbf{15.90}\\

IFT \cite{Lu2022-iv}      
& 16.87\\

TractSeg-LE \cite{Liu2022-oa}        
& 20.41\\

LwF \cite{Lwf}  
& 18.19 \\

PLOP \cite{PLOP} 
&18.94 \\

EWF \cite{EWF}
& 18.26\\

MCLOS \cite{MCLOS}
& 18.46\\

CoinSeg \cite{CoinSeg}
& 19.87\\
\midrule[0.2mm]
\rowcolor{mygray}

\textbf{MultiCo3D (Ours)} 
&  20.33  \\


\bottomrule[0.5mm]
\end{tabular}
\label{tabel:s1}
}
\vspace{5mm}
\end{table}

\clearpage

\begin{table*}[htbp]
\caption{Quantitative comparisons of the SOTA methods on the Preto dataset using DSC (\%). The best results are highlighted in bold.}
\centering
\resizebox{0.99\linewidth}{!}
{
\begin{tabular}{l|ccccccccccccccc}

\toprule[0.5mm]
  
  & \multicolumn{3}{c|}{\textbf{60-12 (2 steps)}}                         & \multicolumn{3}{c|}{\textbf{48-24 (2 steps)}}                         & \multicolumn{3}{c|}{\textbf{36-36 (2 steps)}}   & \multicolumn{3}{c|}{\textbf{24-48 (2 steps)}} & \multicolumn{3}{c}{\textbf{12-60 (2 steps)}}  \\
          
    \multirow{-2}{*}{\textbf{Method}} & 1-60       & 61-72          & \multicolumn{1}{c|}{all}               & 1-48            & 49-72    & \multicolumn{1}{c|}{all}      & 1-36  &37-72  & \multicolumn{1}{c|}{all}   & 1-24    & 25-72              & \multicolumn{1}{c|}{all}            & 1-12& 13-72                 & all               \\ \midrule[0.3mm]


CFT \cite{Lu2022-iv}        
& 63.24 &  49.16 &\multicolumn{1}{c|}{60.89} 
& 60.95 &  44.50 &\multicolumn{1}{c|}{55.47} 
& 60.31 &  51.94 &\multicolumn{1}{c|}{56.13} 
& 61.68 &  41.36 &\multicolumn{1}{c|}{48.13} 
& 64.57 &  38.67 & 42.99\\

IFT \cite{Lu2022-iv}      
& 62.35 &  62.54 &\multicolumn{1}{c|}{62.38} 
& 60.84 &  57.63 &\multicolumn{1}{c|}{59.77} 
& 61.42 &  53.07 &\multicolumn{1}{c|}{ 57.25} 
& 60.34 &  24.51 &\multicolumn{1}{c|}{36.45} 
& 63.51 &  16.92 &24.69 \\

TractSeg-LE \cite{Liu2022-oa}        
&57.98  & 43.81  &\multicolumn{1}{c|}{55.62} 
&55.22  & 45.62  &\multicolumn{1}{c|}{52.02} 
&53.67  & 48.63  &\multicolumn{1}{c|}{51.15} 
&52.73  & 43.30  &\multicolumn{1}{c|}{46.44} 
&55.35  & 39.14  &41.84\\

LwF \cite{Lwf}  
& 65.38 &  57.24 &\multicolumn{1}{c|}{64.02} 
& 65.24 &  55.37 &\multicolumn{1}{c|}{61.95} 
& 64.85 &  55.12 &\multicolumn{1}{c|}{59.99} 
& 63.25 &  53.69 &\multicolumn{1}{c|}{56.88} 
& 64.17 &  52.70 & 54.61\\

PLOP \cite{PLOP} 
& 61.67 &  53.29 &\multicolumn{1}{c|}{60.27} 
& 60.53 &  52.36 &\multicolumn{1}{c|}{57.81} 
& 60.38 &  50.63 &\multicolumn{1}{c|}{55.51} 
& 62.24 &  48.38 &\multicolumn{1}{c|}{53.00} 
& 63.86 &  46.09 & 49.05\\

EWF \cite{EWF}
& 47.31 &   31.64&\multicolumn{1}{c|}{44.70} 
& 46.35 &   31.42&\multicolumn{1}{c|}{41.37} 
& 45.75 &   30.96&\multicolumn{1}{c|}{38.34} 
& 48.63 &   34.91&\multicolumn{1}{c|}{39.48} 
& 47.79 &   36.43&38.32 \\

CoinSeg \cite{CoinSeg}
& 64.38 &  55.38 &\multicolumn{1}{c|}{62.88} 
& 63.79 &  53.27 &\multicolumn{1}{c|}{60.28} 
& 65.31 &  52.03 &\multicolumn{1}{c|}{58.17} 
& 65.47 &  51.06 &\multicolumn{1}{c|}{55.86} 
& 63.89 &  50.28 & 52.55\\
\midrule[0.2mm]
  \rowcolor{mygray}    
\textbf{MultiCo3D (Ours)} 
& \textbf{70.59} & \textbf{65.54}  &\multicolumn{1}{c|}{\textbf{69.75}} 
& \textbf{69.24} & \textbf{63.12}  &\multicolumn{1}{c|}{\textbf{67.20}} 
& \textbf{68.67} & \textbf{61.97}  &\multicolumn{1}{c|}{\textbf{65.29}} 
& \textbf{68.54} & \textbf{60.19}  &\multicolumn{1}{c|}{\textbf{62.97}}
& \textbf{70.23} & \textbf{58.83}  & \textbf{60.73}\\

\bottomrule[0.5mm]
\end{tabular}
\label{tabel:s2}
}
\end{table*}

\begin{table*}[htbp]
\centering
\caption{List of 72 fiber tracts with abbreviations and corresponding full names.}
\resizebox{\linewidth}{!}
{
\begin{tabular}{l|l|l|l}
\toprule[0.5mm]

\textbf{Abbreviation} & \textbf{Full Name} & \textbf{Abbreviation} & \textbf{Full Name} \\ \midrule[0.3mm]
AF\_left              & Arcuate Fascicle (Left)         & AF\_right            & Arcuate Fascicle (Right)       \\ \hline
ATR\_left             & Anterior Thalamic Radiation (Left) & ATR\_right          & Anterior Thalamic Radiation (Right) \\ \hline
CA                    & Commissure Anterior             & CC                   & Corpus Callosum                \\ \hline
CC\_1                 & Corpus Callosum (Rostrum)       & CC\_2                & Corpus Callosum (Genu)         \\ \hline
CC\_3                 & Corpus Callosum (Rostral Body)  & CC\_4                & Corpus Callosum (Anterior Midbody) \\ \hline
CC\_5                 & Corpus Callosum (Posterior Midbody) & CC\_6              & Corpus Callosum (Isthmus)      \\ \hline
CC\_7                 & Corpus Callosum (Splenium)      & CG\_left             & Cingulum (Left)                \\ \hline
CG\_right             & Cingulum (Right)                & CST\_left            & Corticospinal Tract (Left)     \\ \hline
CST\_right            & Corticospinal Tract (Right)     & MLF\_left            & Middle Longitudinal Fascicle (Left) \\ \hline
MLF\_right            & Middle Longitudinal Fascicle (Right) & FPT\_left         & Fronto-Pontine Tract (Left)    \\ \hline
FPT\_right            & Fronto-Pontine Tract (Right)    & FX\_left             & Fornix (Left)                  \\ \hline
FX\_right             & Fornix (Right)                  & ICP\_left            & Inferior Cerebellar Peduncle (Left) \\ \hline
ICP\_right            & Inferior Cerebellar Peduncle (Right) & IFO\_left         & Inferior Occipito-Frontal Fascicle (Left) \\ \hline
IFO\_right            & Inferior Occipito-Frontal Fascicle (Right) & ILF\_left   & Inferior Longitudinal Fascicle (Left) \\ \hline
ILF\_right            & Inferior Longitudinal Fascicle (Right) & MCP             & Middle Cerebellar Peduncle     \\ \hline
OR\_left              & Optic Radiation (Left)          & OR\_right            & Optic Radiation (Right)        \\ \hline
POPT\_left            & Parieto-Occipital Pontine (Left) & POPT\_right          & Parieto-Occipital Pontine (Right) \\ \hline
SCP\_left             & Superior Cerebellar Peduncle (Left) & SCP\_right         & Superior Cerebellar Peduncle (Right) \\ \hline
SLF\_I\_left          & Superior Longitudinal Fascicle I (Left) & SLF\_I\_right  & Superior Longitudinal Fascicle I (Right) \\ \hline
SLF\_II\_left         & Superior Longitudinal Fascicle II (Left) & SLF\_II\_right & Superior Longitudinal Fascicle II (Right) \\ \hline
SLF\_III\_left        & Superior Longitudinal Fascicle III (Left) & SLF\_III\_right & Superior Longitudinal Fascicle III (Right) \\ \hline
STR\_left             & Superior Thalamic Radiation (Left) & STR\_right         & Superior Thalamic Radiation (Right) \\ \hline
UF\_left              & Uncinate Fascicle (Left)         & UF\_right           & Uncinate Fascicle (Right)       \\ \hline
T\_PREF\_left         & Thalamo-Prefrontal (Left)        & T\_PREF\_right       & Thalamo-Prefrontal (Right)      \\ \hline
T\_PREM\_left         & Thalamo-Premotor (Left)          & T\_PREM\_right       & Thalamo-Premotor (Right)        \\ \hline
T\_PREC\_left         & Thalamo-Precentral (Left)        & T\_PREC\_right       & Thalamo-Precentral (Right)      \\ \hline
T\_POSTC\_left        & Thalamo-Postcentral (Left)       & T\_POSTC\_right      & Thalamo-Postcentral (Right)     \\ \hline
T\_PAR\_left          & Thalamo-Parietal (Left)          & T\_PAR\_right        & Thalamo-Parietal (Right)        \\ \hline
T\_OCC\_left          & Thalamo-Occipital (Left)         & T\_OCC\_right        & Thalamo-Occipital (Right)       \\ \hline
ST\_FO\_left          & Striato-Fronto-Orbital (Left)    & ST\_FO\_right        & Striato-Fronto-Orbital (Right)  \\ \hline
ST\_PREF\_left        & Striato-Prefrontal (Left)        & ST\_PREF\_right      & Striato-Prefrontal (Right)      \\ \hline
ST\_PREM\_left        & Striato-Premotor (Left)          & ST\_PREM\_right      & Striato-Premotor (Right)        \\ \hline
ST\_PREC\_left        & Striato-Precentral (Left)        & ST\_PREC\_right      & Striato-Precentral (Right)      \\ \hline
ST\_POSTC\_left       & Striato-Postcentral (Left)       & ST\_POSTC\_right     & Striato-Postcentral (Right)     \\ \hline
ST\_PAR\_left         & Striato-Parietal (Left)          & ST\_PAR\_right       & Striato-Parietal (Right)        \\ \hline
ST\_OCC\_left         & Striato-Occipital (Left)         & ST\_OCC\_right       & Striato-Occipital (Right)       \\ \bottomrule[0.5mm]

\end{tabular}
}
\label{tab:S3}
\end{table*}
\clearpage

{
\clearpage
    \small
    \bibliographystyle{ieeenat_fullname}
    \bibliography{main}
}

\end{document}